\documentclass[11pt]{article}

\usepackage{lmodern} 
\usepackage[T2A]{fontenc}
\usepackage[utf8]{inputenc}
\usepackage[russian,greek,english]{babel}
\usepackage{CJKutf8}

\usepackage[final]{acl}

\usepackage{times}
\usepackage{latexsym}

\usepackage{tipa}

\usepackage{tikz}
\usetikzlibrary{arrows.meta}
\usetikzlibrary{calc}

\usepackage{amsmath}
\usepackage{amsfonts}

\usepackage[T1]{fontenc}

\usepackage{devanagari}

\usepackage[utf8]{inputenc}

\usepackage{microtype}

\usepackage{inconsolata}

\usepackage{graphicx}

\usepackage{tabularx}
\usepackage{booktabs}
\usepackage{multirow}
\usepackage{placeins}

\usepackage{enumitem}

\addto\extrasenglish{%
}

\title{Linear Script Representations in Speech Foundation Models\\ Enable Zero-Shot Transliteration}

\author{
  \textbf{Ryan Soh-Eun Shim\textsuperscript{1,2}},
  \textbf{Kwanghee Choi\textsuperscript{3}},
  \textbf{Kalvin Chang\textsuperscript{5}},
  \textbf{Ming-Hao Hsu\textsuperscript{6}},
  \textbf{Florian Eichin\textsuperscript{1,2}},
  \\
  \textbf{Zhizheng Wu\textsuperscript{6}},
  \textbf{Alane Suhr\textsuperscript{5}},
  \textbf{Michael A. Hedderich\textsuperscript{1,2}},
  \textbf{David Harwath\textsuperscript{3}},
  \textbf{David R. Mortensen\textsuperscript{4}},
  \\
  \textbf{Barbara Plank\textsuperscript{1,2}}
\\
  \textsuperscript{1}LMU Munich, Germany,
  \textsuperscript{2}Munich Center for Machine Learning, Germany,\\
  \textsuperscript{3}University of Texas at Austin, USA,
  \textsuperscript{4}Carnegie Mellon University, USA,\\
  \textsuperscript{5}University of California, Berkeley, USA,
  \textsuperscript{6}Chinese University of Hong Kong, Shenzhen, China
\\
  \small{
    \textbf{Correspondence:} \href{mailto:email@domain}{s.shim@lmu.de}
  }
}

\begin{document}
\maketitle
\begin{abstract}
Multilingual speech foundation models such as Whisper are trained on web-scale data, where data for each language consists of a myriad of regional varieties. However, different regional varieties often employ different scripts to write the same language, rendering speech recognition output also subject to non-determinism in the output script. To mitigate this problem, we show that script is linearly encoded in the activation space of multilingual speech models, and that modifying activations at inference time enables direct control over output script.
We find the addition of such script vectors to activations at test time can induce script change even in unconventional language-script pairings (e.g.\ Italian in Cyrillic and Japanese in Latin script). We apply this approach to inducing post-hoc control over the script of speech recognition output, where we observe competitive performance across all model sizes of Whisper.
\end{abstract}

\section{Introduction}

Multilingual Automatic Speech Recognition (ASR) models such as Whisper \citep{radford2023robust} map speech in different languages to their corresponding written form, where such written language is represented in various scripts, such as Cyrillic, Chinese, or Latin characters.
Similar to how one can translate between languages, it is also possible to write the same spoken word in different scripts via \textit{transliteration}.
For instance, a Russian audio sample of the word meaning \textit{computer}, typically written in Cyrillic script as \foreignlanguage{russian}{компьютер}, can be written in Latin characters as \textit{kompyuter} instead.

\begin{figure}[t]
\centering
\resizebox{0.85\columnwidth}{!}{%
\begin{tikzpicture}[
    >=Stealth,
    font=\sffamily\footnotesize,
    box/.style={
        draw,
        rounded corners=3pt,
        thick,
        minimum width=3.6cm,
        minimum height=1.2cm,
        align=center,
        fill=gray!5
    },
    whisper/.style={
        draw,
        rounded corners=6pt,
        very thick,
        minimum width=3.6cm,
        minimum height=2.0cm,
        fill=blue!6
    },
    arrow/.style={->, very thick},
    magic/.style={->, very thick, purple},
    label/.style={font=\sffamily\scriptsize}
]

\node[box] (audio) at (0,0) {
    \textbf{Audio}\\
    \emph{Italian speech}\\
    \textipa{/bwOn"dZorno/}
};

\node[box] (model) at (0,-2.2) {
    \textbf{Whisper}\\
    activation space
};

\node[box] (native) at (-2.2,-4.6) {
    \textbf{Output}\\
    buongiorno\\
    \textipa{/bwOn"dZorno/}
};

\node[box, fill=red!6] (latin) at (2.2,-4.6) {
    \textbf{Output}\\
    {\foreignlanguage{russian}{буонджорно}}\\
    \textipa{/bwOn"dZorno/}
};

\draw[arrow] (audio) -- (model)
    node[midway, right, label] {encode};

\draw[arrow] (model) -- (native)
    node[midway, left, label] {decode};

\draw[magic] (model) -- (latin)
    node[midway, right, label] {+ script vector};

\end{tikzpicture}%
}
\caption{We induce script change in the transcriptions of Whisper, which is done by identifying ``script directions'' in activation space. Adding such directions to activations at test time induces the desired script. The example shows inducing the Italian word \textit{buongiorno} to be transcribed in Cyrillic characters. Phonetic transcriptions are provided to illustrate the pronunciation of the transliterations.}
\end{figure}
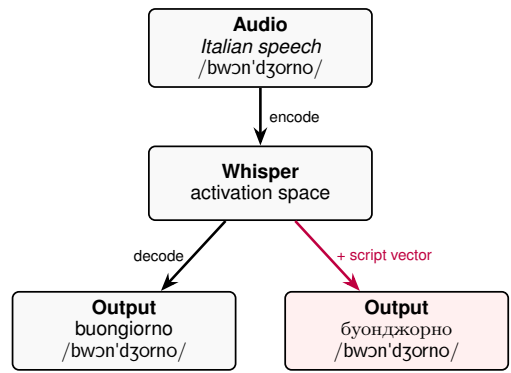

\begin{figure*}[t]
\centering
\resizebox{\textwidth}{!}{%
\begin{tikzpicture}[
  >=Stealth,
  font=\sffamily\footnotesize,
  paneltitle/.style={font=\sffamily\small\bfseries},
  label/.style={font=\sffamily\footnotesize},
  mathlabel/.style={font=\sffamily},
  codelabel/.style={font=\footnotesize}
]

\def\panelSep{6.0}

\begin{scope}[shift={(0,0)}]

\draw[->, gray!50] (-0.3,0) -- (5.2,0);
\draw[->, gray!50] (0,-0.3) -- (0,4.4);

\node[anchor=west, paneltitle] at (0,4.7) {(1) Collect};

\foreach \x/\y in {1.0/1.2, 1.3/0.9, 0.9/1.6, 1.6/1.4, 1.4/1.9}
  \fill[yellow!80!black] (\x,\y) circle (1.8pt);

\foreach \x/\y in {3.6/3.4, 4.0/3.7, 3.8/4.1, 4.3/3.9, 4.0/3.1}
  \fill[blue!70] (\x,\y) circle (1.8pt);

\node[label, yellow!80!black] at (1.3,0.4) {source script};
\node[label, blue!70] at (3.9,2.6) {target script};

\coordinate (srcEx) at (1.6,1.4);
\fill[yellow!80!black] (srcEx) circle (1.8pt);

\coordinate (trgEx) at (4.3,3.9);
\fill[blue!70] (trgEx) circle (1.8pt);

\node[codelabel, anchor=west] at ($(srcEx)+(0.18,0.18)$){\foreignlanguage{russian}{машина}};

\node[codelabel, anchor=west] at ($(trgEx)+(0.18,0.18)$){mashina};

\end{scope}

\begin{scope}[shift={(\panelSep,0)}]

\draw[->, gray!50] (-0.3,0) -- (5.2,0);
\draw[->, gray!50] (0,-0.3) -- (0,4.4);

\node[anchor=west, paneltitle] at (0,4.7) {(2) Isolate};

\foreach \x/\y in {1.0/1.2, 1.3/0.9, 0.9/1.6, 1.6/1.4, 1.4/1.9}
  \fill[yellow!40!black] (\x,\y) circle (1.4pt);

\foreach \x/\y in {3.6/3.4, 4.0/3.7, 3.8/4.1, 4.3/3.9, 4.0/3.1}
  \fill[blue!40] (\x,\y) circle (1.4pt);

\coordinate (muS) at (1.24,1.4);
\coordinate (muT) at (3.94,3.64);

\fill[yellow!80!black] (muS) circle (2.6pt);
\fill[blue!70] (muT) circle (2.6pt);

\node[below left, mathlabel] at (muS) {$\mu_{\mathrm{src}}$};
\node[above right, mathlabel] at (muT) {$\mu_{\mathrm{trg}}$};

\draw[->, very thick, purple]
  (muS) -- (muT)
  node[midway, above left, label] {script direction};

\end{scope}

\begin{scope}[shift={(2*\panelSep,0)}]

\draw[->, gray!50] (-0.3,0) -- (5.2,0);
\draw[->, gray!50] (0,-0.3) -- (0,4.4);

\node[anchor=west, paneltitle] at (0,4.7) {(3) Add};

\coordinate (test) at (2.0,1.5);
\fill[black] (test) circle (2.2pt);

\node[below left, label] at (test)
  {\foreignlanguage{russian}{компьютер}};

\draw[->, thick, purple]
  (test) -- +(2.7,2.2);

\coordinate (steered) at (4.7,3.7);
\fill[red!70] (steered) circle (2.2pt);

\node[above right] at (steered) {kompyuter};

\end{scope}

\end{tikzpicture}%
}
\caption{Illustration of our method for extracting script vectors. (1) For each decoder layer, we collect activations in the source (yellow) and target (blue) script. (2) We isolate a script direction by subtracting the mean of the target activations from the mean of the source activations for each layer. (3) At test time, we add the script direction to the activations to induce the transcription to be in the target script.
}
\label{fig:steering_methodology}
\end{figure*}
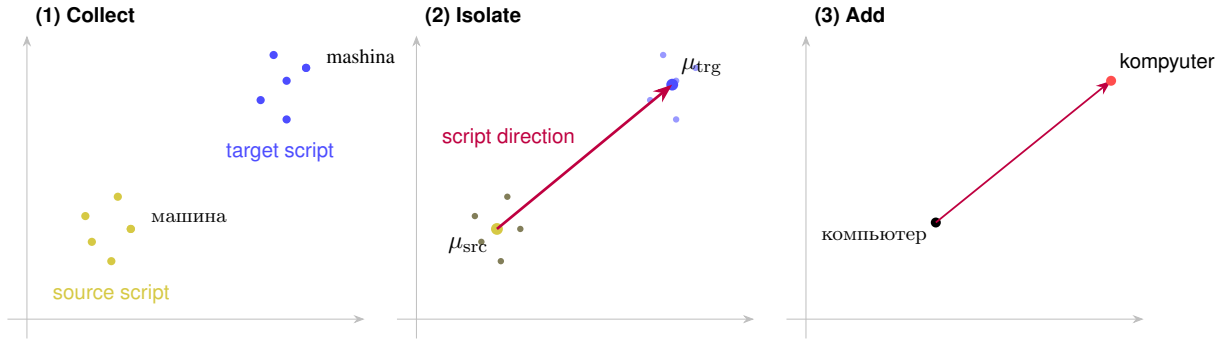

Moreover, many languages are written in multiple scripts. For example, the pluricentric language of Serbo-Croatian encompasses multiple mutually-intelligible varieties and multiple orthographic standards. Within Serbo-Croatian, Serbian, for instance, is written in both Cyrillic and Latin characters, while Croatian leans heavily towards the usage of only the Latin script. Similarly, Mandarin is written in Traditional and Simplified Chinese characters, where the former is primarily used in Taiwan and the latter in China. Thus models trained on web-scale data have been trained on speech within the same language that has transcriptions in different scripts, leading to non-determinism. As scripts carry sociopolitical connotations, finer control over the output script may help better adapt the output to different scripts (e.g.\ Latin script in Croatian contexts \citep{jovanovic_assertive_2018}; Traditional Chinese in Taiwanese contexts \citep{su2021chineseness}). As such, post-hoc control over the script of the output is desirable.

For inducing post-hoc control, an increasing body of work in text-based models has found high-level concepts such as refusal \citep{arditi2024refusal,wang2025refusal} and truthfulness \citep{li2024inferencetimeinterventionelicitingtruthful, marks2024the} to be represented as a linear direction in the activation space of neural models \cite{lrp}. Such findings have been used to induce test-time interventions that steer generation output to the specified concept direction with simple vector arithmetic, where directions can be added to activations at test time to induce behavior \citep{turner2023activation}. In this work, we leverage this strategy \emph{to induce script change in speech foundation models} by way of such steering. Our findings are as follows: 

\begin{enumerate}[label={}, leftmargin=*]
    \item \textbf{Script confusion hinders control over ASR output.}
    We propose script vectors as a way to address the problem of \textit{script confusion}, where languages written in multiple scripts suffer from lack of control over the script of speech recognition outputs.
    \item \textbf{Script is a linear direction in activation space.} \\We find \textit{script} to be represented linearly in multilingual ASR models, enabling post-hoc control over the script of the ASR output by way of adding script vectors to the hidden states. We show that such an approach can induce transliterations in novel directions (e.g.\ Italian in Cyrillic characters) and mitigate script confusion without any training, requiring as few as 1 good quality example pair.

    \item \textbf{In-depth analysis of script vectors.} We perform an in-depth analysis of the layer-wise presence and linguistic characteristics of transliterations induced with script vectors, where we find script to be linearly represented in all decoder layers, and that the induced transliterations tend to be sound-based, reflecting actual pronunciation.
    
\end{enumerate}

\begin{table*}[t]
\small
\centering
\begin{tabular}{p{1.5cm} p{4cm} p{4cm} p{4cm}}
\hline
\textbf{Language} & \textbf{Predicted romanization (ours)} & \textbf{Deterministic romanization} & \textbf{Original transcript} \\
\hline
Russian
& Del Patro imel rane \textbf{premuscestvo} vo vtorom seti, no \textbf{eta} takzhe potrebvala taj brejka posle dostizzenia 6-6. & del potro imel rannee \textbf{preimushchestvo} vo vtorom sete no \textbf{eto} takzhe potrebovalo tai-breika posle dostizheniya 6-6 & \foreignlanguage{russian}{дель потро имел раннее преимущество во втором сете но это также потребовало тай-брейка после достижения 6-6}\\
\addlinespace[1em]
Hindi
& Kuch anvon me aar sthir kendrak hota hai jiska matlab yaha ki \textbf{unme} thode ya bina kisi jhatke se \textbf{tutne} ki \textbf{pravati} hoti hai.
& kuch annuom mem asthir kemdrak hotaa hai jisakaa matalab yah hai ki \textbf{unamem} thodde yaa binaa kisii jhattake se \textbf{ttuuttane} kii \textbf{pravrtti} hotii hai & {\dn kuchh aNuom\~meM asthir keMdrak hotA hai jiskA matalab yah hai ki unameM thoRe yA binA kisI jhaTake se TUTane kI pravRtti hotI hai}\\
\addlinespace[1em]
Greek 
& \textbf{Skeftite} \textbf{ti} \textbf{diadromi} me ski os mia \textbf{omia} \textbf{kalipsi} \textbf{apostasis} pezoporondas & \textbf{skefteite} \textbf{ten} \textbf{diadrome} me ski os mia \textbf{omoia} \textbf{kalypse} \textbf{apostases} pezoporondas & \foreignlanguage{greek}{σκεφτείτε την διαδρομή με σκι ως μια όμοια κάλυψη απόστασης πεζοπορώντας} \\
\addlinespace[1em]
Japanese
& Hong Kong no chiheisen wo egaite tachinarabu biru-gun \textbf{wa}, victoria harbour no mizu no sonzai ni yote kirabiyakana bougurafu ni tatoerarete imasu.
& honkon no chiheisen wo egai te tachi narabu biru gun \textbf{ha} bikutoria haabaa no mizu no sonzai niyotte kirabiyakana bou gurafu ni tatoe rareteimasu & \begin{CJK}{UTF8}{min}
香港の地平線を描いて立ち並ぶビル群は ビクトリア ハーバーの水の存在によって きらびやかな棒グラフに例えられています \end{CJK}\\
\addlinespace[1em]
Korean
& Internet\begin{CJK}{UTF8}{mj}은 집\end{CJK}tanjeogin communicationgwa saramgwa saramgane communicationyosoreul modu \textbf{gachugo itda}. & inteoneseun jibdanjeogin keomyunikeisyeongwa saramgwa saram ganyi keomyunikeisyeon yosoreul modu \textbf{gajcugo issda} & \begin{CJK}{UTF8}{mj} 인터넷은 집단적인 커뮤니케이션과 사람과 사람 간의 커뮤니케이션 요소를 모두 갖추고 있다 \end{CJK} \\

\hline
\end{tabular}
\caption{Sample predictions of our romanization with Whisper large-v2. Parts where deterministic romanization output does not reflect actual pronunciation but our prediction does are typeset in bold.}
\label{tab:roman_examples}
\end{table*}

\section{Methodology}
\label{sec:methodology}

We seek to isolate specific script directions in a speech foundation model's activation space with \textit{script vectors} such that the addition of the found script vector to the activations can control the script of the generated text at test time.
In this work, we focus on encoder-decoder speech foundation models, where the encoder processes speech and the decoder is an auto-regressive language model that processes text. %

\subsection{Script Steering}\label{chapter:script_steering}
We obtain one steering vector per decoder layer by computing the difference in mean activations \citep{arditi2024refusal, marks2024the} between samples of two given scripts. Our methodology consists of three steps: activation collection and filtering, direction isolation, and activation addition, where \autoref{fig:steering_methodology} provides an overview.

\paragraph{Activation collection}
\label{sec:act_col}
Let $D$ be the hidden size and $L$ the number of decoder layers. For each audio example, we decode two transcriptions with different text prompts (fed into the decoder): a \emph{source} prompt $p^{SRC}$ that contextually biases the transcription towards the source script and a \emph{target} prompt $p^{TRG}$ that biases towards the target script (where the scripts are in the same language for this work).\footnote{We provide the prompts we use in \autoref{sec:appendix}.} For a text sequence $z$ decoded from the audio, let $\mathbf{x}_{\ell}(p^{SRC}, z)_t\in\mathbb{R}^D$ be the activation (output) of decoder layer $\ell$ at token position $t$ given source prompt $p^{SRC}$, and $\mathbf{x}_{\ell}(p^{TRG}, z)_t\in\mathbb{R}^D$, the corresponding activation under the target prompt $p^{TRG}$. For each layer $\ell$, we average activations per example via a mean pool over the token positions:

\begin{equation}
\bar{\mathbf{x}}_{\ell}(p, z)
:= \frac{1}{|z|} \sum_{t \in z}\mathbf{x}_{\ell}(p, z)_t%
\label{eq:pooling}
\end{equation}

Given a training set $\mathcal{D}$, we then compute the mean layer-wise activations using the source prompt $\mathcal{D}_{\text{SRC}} := \{\bar{\mathbf{x}}_{\ell}(p^{SRC}, z) | z \in \mathcal{D} \}$ and the target prompt $\mathcal{D}_{\text{TRG}} := \{\bar{\mathbf{x}}_{\ell}(p^{TRG}, z) | z \in \mathcal{D} \}$, respectively:

\begin{equation}
\begin{split}
\mathbf{v}^{\text{SRC}}_{\ell}
&:= \frac{1}{|\mathcal{D}_{\text{SRC}}|}
   \sum_{\bar{\mathbf{x}}\in\mathcal{D}_{\text{SRC}}}\bar{\mathbf{x}}, \\
\mathbf{v}^{\text{TRG}}_{\ell}
&:= \frac{1}{|\mathcal{D}_{\text{TRG}}|}
   \sum_{\bar{\mathbf{x}}\in\mathcal{D}_{\text{TRG}}}\bar{\mathbf{x}}.
\end{split}
\label{eq:dom}
\end{equation}

To avoid noise due to unsuccessful prompting (\textit{i.e.}, the model outputting the non-target script), we additionally apply filtering while collecting and averaging activations. We retain activations where the transcriptions are likely to be in the target script, which results in cleaner activations from which the vector can be isolated.
We use the normalized Levenshtein edit distance $\bar{d}_\text{edit}$ between the transcription prediction $z$ and the ground truth $\hat{z}$:
\begin{equation}
\label{eq:script-similarity}
\bar{d}_\text{edit}(z, \hat{z}) := \frac{d_\text{edit}(z, \hat{z})}{\mathrm{max(|z|, |\hat{z}|)}}.
\end{equation}
Based on the threshold hyperparameter $\theta \in \mathbb{R}$, we only keep the activations with $\bar{d}_\text{edit} < \theta$ within $\mathcal{D}_\text{SRC}$ and $\mathcal{D}_\text{TRG}$.

\paragraph{Direction isolation}
Based on the linear representation hypothesis \cite{lrp} where high-level concepts are represented as directions in activation space, we use $\mathbf{v}^{\text{SRC}}_{\ell}$ and $\mathbf{v}^{\text{TRG}}_{\ell}$ to obtain the \emph{steering vector} $\mathbf{r}_{\ell}$ for layer $\ell$:
\begin{align}
    \mathbf{r}_{\ell} \;:=\; \mathbf{v}^{\text{SRC}}_{\ell} - \mathbf{v}^{\text{TRG}}_{\ell}\;\in\mathbb{R}^D.
\end{align}

\paragraph{Activation addition}
During transcription, at each step of decoding,  we edit the \emph{current last-token} activation $\mathbf{h}_{t,\ell}\in\mathbb{R}^{D}$ at layer $\ell$ by adding the script vector to the activation:
\begin{equation}
\mathbf{h}_{t,\ell}^{\text{steered}} \;:=\; \mathbf{h}_{t,\ell} \;+\; \sigma\,\mathbf{r}_{\ell},
\label{eq:residual-steer}
\end{equation}
where $\sigma$ controls the strength of the direction. In our experiments, we follow \citet{wang-etal-2025-bridging} in steering all layers at once, using the same sigma value for all layers. We show in \autoref{sec:probing} that steering all layers is arguably preferable, due to script being linearly represented in all layers. In our experiments, we tune sigma on a validation set, the details of which are described in \autoref{sec:dataset}.

\section{Experiments}

We evaluate our setup on two use cases: (1) mitigating script confusion, where multiple scripts occur within a single language's speech recognition output, and (2) transliteration, where we transcribe speech from a source language in the orthography (script) of a target language, where the target script is not associated with the source language (e.g.\ transcribing Italian in Cyrillic script). We focus our transliteration experiments on romanization and cyrillization, due to the prevalence of languages written in these two scripts. We provide further details below. %

\subsection{Script confusion}

\begin{figure*}
  \centering
  \includegraphics[width=\textwidth]{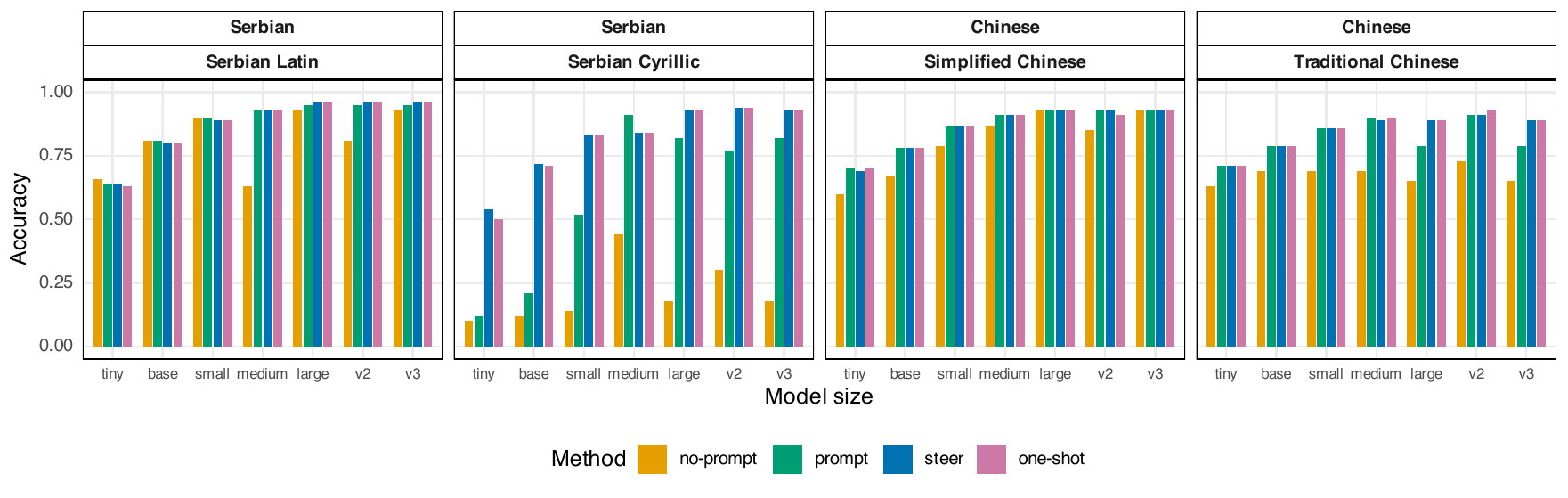}
  \caption{Script confusion mitigation accuracy. Line color shows method. The x-axis shows model size, while the y-axis shows normalized edit similarity to ground truth in target script (\autoref{eq:script-similarity}).}
  \label{fig:script_confusion}
\end{figure*}

\label{sec:script_confusion}
The goal of the script confusion setting is twofold: to measure the extent to which transcriptions are unstable in script, and whether our script steering mitigates such an instability.
We focus our experiments on two languages that are written in multiple scripts: Serbian (written in Latin and Cyrillic script) and Mandarin (written in Traditional and Simplified Chinese). For generating ground truths for these script pairs, we employ deterministic, language-specific tools: cyrtranslit \citep{georges_labreche_nov2025} for Serbian and OpenCC \citep{kuo2024opencc} for Mandarin.
We compare the following approaches:
\paragraph{no-prompt}
We empirically measure the likelihood of speech recognition being in a given script, by directly running the model without any modifications, and measuring its output against the gold transcription in the target script. This provides a notion of how much speech recognition in a given language is likely to be unstable in the script. In this setup, the only input is the audio itself.

\paragraph{prompt}
For our primary baseline, we measure the performance of contextually biasing the output script by prepending the decoder with text in the target script.\footnote{Based on: \href{https://github.com/openai/whisper/discussions/277}{https://github.com/openai/whisper/discussions/277}}

\paragraph{steer (ours)}
Finally, we measure the extent to which our proposed activation steering method (\autoref{sec:methodology}) can mitigate script confusion, where we isolate directions in activation space that encode script, and add them to activations at test time. The same prompts as the \textbf{prompt} baseline are used to obtain script vectors. In addition, we include results of learning script vectors using a single example, which we label as \textbf{one-shot}. We describe the one-shot setup in more detail in \autoref{sec:one_shot}.

\subsection{Transliteration}
\label{sec:transliteration}

In the transliteration setting, we evaluate romanization and cyrillization, for which deterministic tools are available \citep{hermjakob-etal-2018-box, georges_labreche_nov2025}. For romanization, we evaluate on five languages with diverse scripts that are supported by the tools uroman \citep{hermjakob-etal-2018-box} and pykakasi \citep{miura2024pykakasi}, which include Japanese, Korean, Russian, Greek, and Hindi. For cyrillization, we rely on International Components for Unicode\footnote{\href{https://github.com/unicode-org/icu}{https://github.com/unicode-org/icu}} (ICU) to transliterate the same languages as above into Cyrillic, but instead of Russian we evaluate on Italian, as the former is already in Cyrillic. For source-target script pairs not directly implemented in ICU, we use the deterministic romanization tools above to generate romanization then pass through ICU to generate Cyrillic. 

In addition to \textbf{no-prompt} and \textbf{prompt} from section \ref{sec:script_confusion}, we evaluate steering with script vectors in two different ways:
\paragraph{zero-shot (ours)}
We empirically found that script vectors obtained for a specific source-target pair of scripts generalize zero-shot to other languages and scripts to a reasonable extent. As such, we measure the extent to which the romanization and cyrillization directions (derived from one representative language) when applied to other languages successfully induces the corresponding scripts.

\paragraph{pseudo label (ours)}
As the zero-shot approach has shown to induce transliterations with varying degrees of success, we hypothesize that the transliterations induced in such a manner can be used as pseudo transcriptions to learn script vectors better suited for the new language. As such, we repeat the script steering procedure using the steered model to induce script vectors, where we collect activations on the steered model and learn a script vector on these newly collected activations.

\subsection{Evaluation}
The goal of both settings is to transcribe speech using the correct target script.
To quantify the degree to which the predicted text matches the ground truth, we compute the edit distance of our transcription against such ground truth at the character level, normalized by the length of the reference or hypothesis (whichever is longer).
We then subtract the resulting value from 1 to convert the distance into a similarity metric for ease of interpretation (\autoref{eq:script-similarity}).
In our results, we term this \textit{accuracy}, where 1.0 is text that fully matches the ground truth. For preprocessing, as our goal is to quantify the degree to which correct transliteration occurs, we remove punctuation, spacing, and portions of the text that are not in the target script.%

\subsection{Dataset and Hyperparameters}
\label{sec:dataset}
We perform our experiments in both settings with the FLEURS \citep{fleurs} dataset, a dataset with parallel speech and text transcriptions in 100+ languages.\footnote{We include dataset statistics in \autoref{tab:data_stats}.}
For each source and target script, we collect activations on 10 samples respectively from the train set, where the first 10 audio samples that exceed the $\theta$ threshold in both source and target scripts are collected. For $\theta$ that serves as the threshold for filtering activations, we use $\theta=0.4$ for script confusion and romanization. For cyrillization, due to lower recognition performance than romanization, we increase to $\theta=0.8$. For $\sigma$ that controls the strength of the script vector, we perform a grid search over $\sigma \in \{0.1, 0.2, 0.3, 0.4, 0.5\}$ on the validation set in FLEURS. We report our results on the test sets of each language using the best $\sigma$.\footnote{For script confusion, best here means the sigma resulting in the highest mean accuracy. For transliteration, due to zero-shot accuracy being unevenly distributed over samples, we use the sigma resulting in the single highest (i.e.\ max) accuracy. For script confusion, the best $\sigma$ under all settings is 0.1. For transliteration, we report $\sigma$ in \autoref{tab:sigmas}.}

As for the speech foundation model, we leverage Whisper \citep{radford2023robust}, where we evaluate our approach on all model sizes: tiny, base, small, medium, large, v2, and v3. for the script confusion setting. For transliteration, we focus on large-v2, the best performing model size in script confusion, which prior work has also found to perform the best across all model sizes \citep{ma-etal-2025-cross}.

\section{Results}
We next present results for our two key setups.
\subsection{Script Confusion}

\paragraph{Script vectors induce the correct target script in smaller models even when prompting does not.}
\autoref{fig:script_confusion} shows our results on mitigating script confusion. We observe that the effectiveness of prompting for controlling the output script tends to be more limited in smaller models, but improves with the size of the model. We see this particularly for inducing Cyrillic in Serbian. For instance, in Whisper tiny, prompting only slightly induces more Cyrillic script than running inference directly, whereas activation steering significantly increases Cyrillic script accuracy.

\paragraph{Script confusion mitigation effectively induces the less frequent scripts of a language.} \label{sec:less_freq_script}
For both Chinese and Serbian, we observe that prompting and steering both induce improvements over the no-prompt baseline primarily for the less frequent (and thus lower resource) script of a given language (\textit{i.e.}, Serbian in Cyrillic and Traditional Chinese). For Serbian in Latin script and Simplified Chinese, the improvements are less pronounced, due to inference already producing higher proportions of transcriptions in these scripts.

\subsection{Transliteration}\label{sec:results-transliteration}

\begin{figure}[t]
  \centering
  \includegraphics[width=0.475\textwidth]{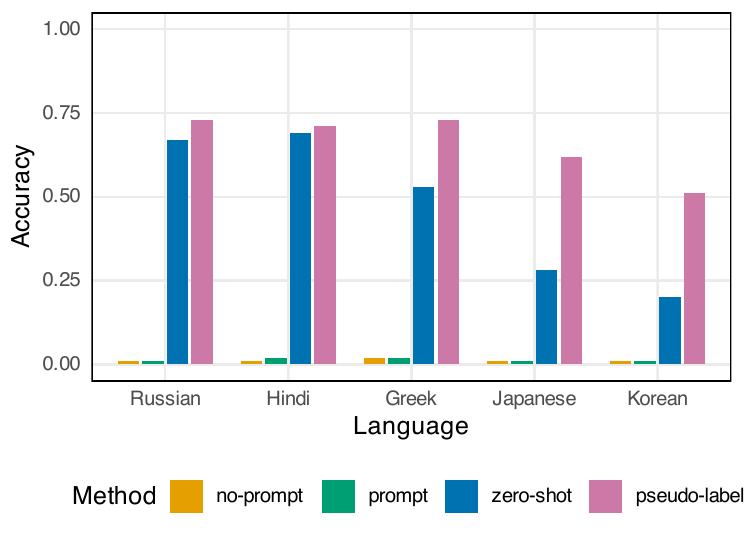}
  \caption{Romanization accuracy across different languages for Whisper large-v2. The x-axis shows languages, while the y-axis shows normalized edit similarity against ground truth in target script (\autoref{eq:script-similarity}).}
  \label{fig:romanization}
\end{figure}

\begin{figure}[t]
  \centering
  \includegraphics[width=0.475\textwidth]{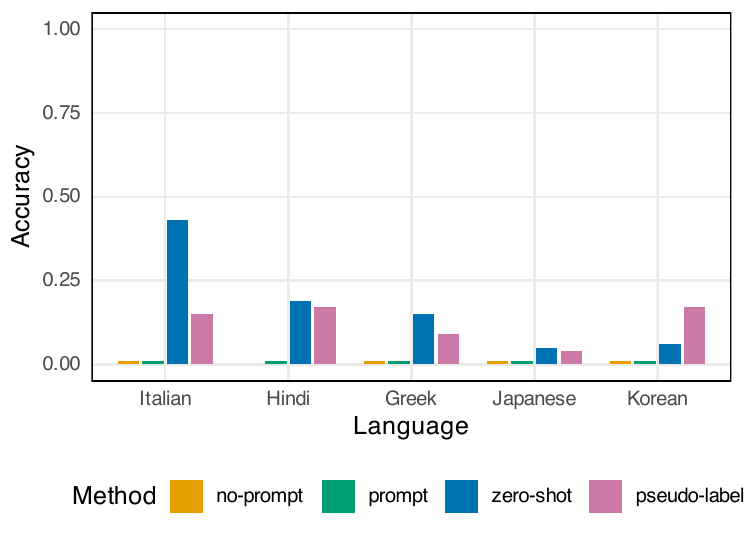}
  \caption{Cyrillization accuracy across different languages for Whisper large-v2. The x-axis shows languages, while the y-axis shows normalized edit similarity against ground truth in target script (\autoref{eq:script-similarity}).}
  \label{fig:cyrilization}
\end{figure}

\paragraph{Script vectors can generalize across languages.}
\label{sec:romanization}
\autoref{fig:romanization} shows the results of directly applying our romanization vector derived from Serbian on different languages, along with results from using the induced transcriptions as pseudo labels to learn a direction adapted to the specific language. We observe that for Hindi and Russian, zero-shot performance is already at 69 and 67\%, respectively, suggesting romanization direction to be broadly similar across languages. The further adaptation step with pseudo labels boosts performance across all five languages, further supporting the hypothesis of similarity between romanization direction across languages.

\paragraph{Script vectors can induce transliterations in unconventional language-script pairs.}
\autoref{fig:cyrilization} shows our results for cyrillization. In comparison with romanization, cyrillization observes more limited performance across most languages, with the exception of Italian, which reaches 43\% accuracy. We provide examples of our Cyrillic transliterations in \autoref{tab:cyril_examples}. We hypothesize the exceptional performance of Italian in cyrillization to be due to it being written in the high-resource Latin script, which the majority of Whisper's training data is in \citep{radford2023robust}. This is also empirically supported by the general success of romanization for all source languages examined in \autoref{sec:romanization}.

\begin{table*}
\centering
\small
\begin{tabular}{p{1.5cm} p{4cm} p{4cm} p{4cm}}
\textbf{Language} &
\textbf{Prediction (ours)} &
\textbf{Deterministic} &
\textbf{Original} \\
\hline
Italian & \foreignlanguage{russian}{Нелло специфико си состене ке е посибиле скоприре су уна персона ста ментэндо интерпретандо ле микроэспрессиони ин модо коррѕтто.} & \foreignlanguage{russian}{нелло специфицо си состиене цхе е поссибиле сцоприре се уна персона ста ментендо интерпретандо ле мицроеспрессиони ин модо цорретто} & nello specifico si sostiene che è possibile scoprire se una persona sta mentendo interpretando le microespressioni in modo corretto\\

 \hline
Hindi & \foreignlanguage{russian}{Вахан коу уси дин такребен 1200 ге МТ ке саме паре дургатана стал се дур лејаја геа.} & \foreignlanguage{russian}{ваахан ко усии дин такариибан 1200 гмт ке самаы пар дургхаттанаастхал се дуур ле йааыаа гаыаа} & {\dn vAhan ko usI din takrIban 1200 gmt ke samay par durghaTnAsthal se dUr le jAyA gayA}\\

 \hline
Greek & \foreignlanguage{greek}{Το} \foreignlanguage{russian}{«Теолака Жинк»} \foreignlanguage{greek}{γήιθκε} \foreignlanguage{greek}{στο} \foreignlanguage{greek}{τραγούδι των} \foreignlanguage{russian}{Бадзан.} & \foreignlanguage{russian}{тоте о лакка синнг егетхеке сто трагоуди тон батзан}  & \foreignlanguage{greek}{τότε ο λάκκα σινγκ ηγήθηκε στο τραγούδι των μπατζάν}\\

 \hline
Japanese & \begin{CJK}{UTF8}{min}ワイルドカードを購入するとお\end{CJK}\foreignlanguage{russian}{токуна бае га аримасу. Минами Африканоу омонаа коујн матауа Минами Африканоу сваутеу на коујну ни нужоу дъјуудщу.} & \foreignlanguage{russian}{уаирудокаадо уо коуныуу сурутоо ена бааи гааримасу   минами афурика но омона коуен матаха минами афурика носубетено кокуритсукоуен ни ныууйоу декимасу} & \begin{CJK}{UTF8}{min}ワイルドカードを購入するとお得な場合があります 南アフリカの主な公園または南アフリカのすべての国立公園に入場できま\end{CJK}\\

\hline
Korean & \begin{CJK}{UTF8}{mj} 그해에 가장 규모가 큰 토너먼트 경기는 12월에 \foreignlanguage{russian}{лас Канитасее фолоу генгианг изо ерлимьимида.} \end{CJK} & \foreignlanguage{russian}{геу хаеыи гайанг гыумога кеун тонеомеонтеу гыеонггинеун 12уеоле расеу канитасеуыи полро гыеонггийангесео ыеолрибнида} & \begin{CJK}{UTF8}{mj}그 해의 가장 규모가 큰 토너먼트 경기는 12월에 라스 카니타스의 폴로 경기장에서 열립니다\end{CJK}\\

 \hline

\end{tabular}
\caption{Sample predictions of our cyrillization with Whisper large-v2. Italian observes the strongest performance, although other languages also observe sensible cyrillization in parts.}
\label{tab:cyril_examples}
\end{table*}

\section{Analysis}
In this section, we present further analyses on our methodology and results.
\paragraph{Script vectors can be learned from one example.}\label{sec:one_shot} %
As our method does not rely on any finetuning, we hypothesize that few high quality activation pairs should already be able to approximate the script direction--and find one pair is sufficient. As such, we repeat our script confusion experiments using a single sample, where we select the first example that passes our filtering threshold described in \autoref{sec:act_col}. \autoref{fig:script_confusion} shows our results. We observe that one sample is already enough to induce script vectors, making our steering approach very sample-efficient. This potentially implies that cyrillization performance in the pseudo-label setup has potential for  improvement, if one isolates the script vector on a single high quality example, as opposed to the current few-shot setup that potentially introduces noise with a larger number of samples.

\paragraph{Script is linearly separable across all decoder layers.}\label{sec:probing}

We analyze the layer-wise presence of script-related information using \citet{marks2024the}'s linear probe. Let $\mathbf{r}_l \in \mathbb{R}^D$ be the script vector for layer $\ell$ as defined in \autoref{chapter:script_steering} and $\mathbf{x}_\ell \in \mathbb{R}^D$ an activation in the same layer. The probe is then $\sigma(\mathbf{r}_l^T \mathbf{x})$, where $\sigma$ is the sigmoid function.
Using the train split of FLEURS, we collect 50 activations per script for Serbian in Cyrillic and Latin script, and Mandarin in Traditional and Simplified script, yielding 100 activations in total for each language.\footnote{We use $\theta = 0.1$ here to ensure the cleanliness of the labels.} On the FLEURS test split, we perform the same collection procedure and subtract the mean of the training split activations as a preprocessing step to center each activation. The probe then predicts the script (e.g., Simplified or Traditional Chinese) and measures accuracy.

\begin{figure}
  \centering
  \includegraphics[width=0.45\textwidth]{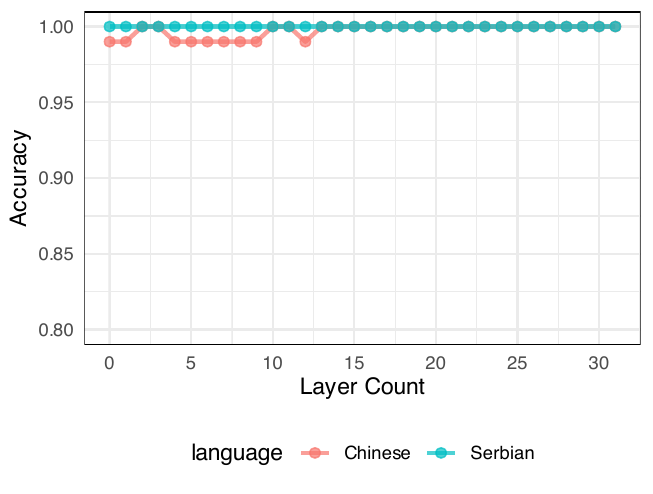}
  \vspace{-1em}
  \caption{Script probing accuracy for Whisper large-v2 on Chinese and Serbian (\autoref{sec:probing}).}
  \label{fig:probe_acc}
\end{figure}

\autoref{fig:probe_acc} shows our results. We observe that for both Serbian and Mandarin, the scripts produced are separable across all layers. One implication of this finding is that, as already in the first decoder layer it is predictable what script a given sample will decode to, we hypothesize one can rely on such probes to selectively apply either prompting or activation steering, instead of waiting until the full forward pass to produce the output in practical scenarios. As we observe in \autoref{sec:less_freq_script} that steering occasionally underperforms prompting for the more frequent script of a given language, this would potentially yield further performance gains. Another implication is that, since script is linearly separable in all layers, it is intuitive then to also apply steering in all layers, instead of selecting only one layer.

\paragraph{Script vectors induce sound-based transliterations.}
In \autoref{tab:roman_examples}, we observe that in comparison with the deterministic transliteration outputs, our prediction appears to reflect the actual pronunciation in many cases. We put in bold parts where the deterministic transliteration is induced by character mappings but does not actually reflect the pronunciation, where steering does. Taking as example the case of Korean, a deterministic transliteration based on character mapping renders the output \textit{gajcugo issda} for the Korean tokens \begin{CJK}{UTF8}{mj} 갖추고 있다\end{CJK}, deviating significantly from the pronunciation \textipa{/ka\t{tC}\super hugoit\textcorner ta/}, which the steered transliteration \textit{gachugo itda} arguably reflects better. Similarly, for the Greek example, modern Greek has gone through vowel mergers that cause \foreignlanguage{greek}{η, ι, υ, ει, οι} to be pronounced as /i/. The original pronunciation in earlier stages of Greek is preserved in the orthography, which the deterministic system relies on for mapping, but is not reflective of how they are actually pronounced in modern Greek \citep{holton2012greek}. In contrast, our steering results consistently render these as \textit{i}, in line with the pronunciation. Similar examples are also found in Hindi and Russian where orthographic cues mislead the deterministic system. Finally, in the case of Japanese, the character \begin{CJK}{UTF8}{min}は\end{CJK} is normally pronounced as \textit{ha} as the deterministic system transliterates, but in the example above serves as a topic particle, in which case the correct pronunciation would be \textit{wa} \citep{martin2003reference}. Our steering correctly reflects this. %

\paragraph{Why are some named entities transcribed in English?}
In \autoref{tab:roman_examples}, we observe that some named entities such as \textit{victoria harbour} and \textit{Hong Kong} are written directly in their English forms, as opposed to being transcribed phonetically the way they are pronounced in the original language (\textit{bikutoria haabaa} and \textit{hon kon} in Japanese). We also observe a related phenomenon  where named entities tend to be the most easily steerable portions of a given transcription when applying smaller sigma values. We hypothesize this phenomenon to be due to some named entities being cross-lingually similar to English in pronunciation, such that they share stronger cross-script alignment. %
As the majority of Whisper's training data is in English \cite{radford2023robust}, it is likely that this strongly affects the representation of such named entities to also be in English for Latin script languages. This is further evidenced by terms such as \textit{Internet} and \textit{communication} in the Korean transliteration in \autoref{tab:roman_examples}, which are loanwords in Korean that are phonetically similar to their English sources. We hypothesize that using data with many named entities or cross-lingually similar sounding words can ease the process of extracting script vectors. %

\section{Related Work}

\paragraph{Script confusion}
There is significant work on the related problem of \textit{language confusion} wherein language models \citep{marchisio-etal-2024-understanding} or code-switched ASR \citep{liu2023reducing} struggles to output the correct language.
In contrast, we focus on script confusion in speech foundation models, which is the problem of not just outputting the correct language but also the correct script.

\citet{omnilingual2025omnilingual} tackled script confusion in multilingual ASR by modifying the language token to include the script.
This approach may require retraining or finetuning for models already trained without script tokens like Whisper. Furthermore, as script is hard-coded in the language tokens, it is not obvious whether transliteration in unseen directions can be induced. %
In contrast, we provide an inference-time, backpropagation-free mechanism for script control that allows transliteration even in unconventional language-script pairs.

\paragraph{Transliteration in NLP and ASR}

Work on transliteration has primarily focused on the text domain for use cases such as Latin script keyboards.
This involves training task-specific models to perform transliteration \citep{ryskina-etal-2020-phonetic,roark2020processing} and finetuning LLMs \citep{purkayastha-etal-2023-romanization,kirov2024context}. 
Within multilingual NLP and ASR, romanization (transliteration into Latin scripts) has provided a crosslingual phonetic space, enabling ``zero-shot'' adaptation to new languages \citep{liu2025transliterations,jaavid2024romansetu,saji2025romanlens,jung2025happiness,lee2025lama,zhu2024transliterated,yan2023towards,nigatu2025exploring}.
None has explored activation steering for transliteration.

\paragraph{Activation steering}

Activation steering is a technique for controlling the outputs of a language model at inference time by extracting steering vectors and adding them to the language model's activations \citep{turner2023steering}.
This method can control model outputs' topic, sentiment, toxicity \citep{turner2023steering}, truthfulness \citep{li2024inferencetimeinterventionelicitingtruthful,marks2024the,ravfogel2025emergence} and refusal of safety-related queries \citep{wang2025refusal,marshall2024refusal}.
This works because many of the aforementioned concepts can be linearly represented in an embedding space \citep{lrp,modell2025origins} and thus controlled with a single vector.
Such methods can be used in speech language models as well \citep{lin2025adaptive,xie2025emosteer,lin2025sarsteer}.
No prior work on activation steering, however, has focused on test-time control of subword properties like sounds, much less orthography (script).

\section{Conclusion and Future Work}
In this work, we show that script information is linearly represented in the activation space of speech foundation models. Such directions enable control over the script of the transcription output and generalize cross-lingually, enabling transliteration in novel directions. One implication of our work is that speech foundation models may already acquire cross-script alignment even without an explicit transliteration step, as recent work has shown transliterating all scripts to a unified script to help speech recognition performance \citep{lee2025lama}. Future work may therefore look into the training dynamics of speech foundation models to investigate how such an alignment is acquired throughout training.

\section*{Limitations}

We acknowledge that there exist multiple possible transliterations for a given word: ones that lean more semantic \citep{li-etal-2007-semantic} and ones that are more phonetically faithful, among other criteria \citep{demirsahin-etal-2022-criteria}.
In this work, we focused only on phonetically faithful transliterations. Furthermore, while we tested various models in preliminary experiments, our experiments focus mostly on the Whisper series due to their widespread usage and influence \citep{ma-etal-2025-cross, peng2025owsmv4improvingopen}. Future work can analyze the training dynamics of open-data speech foundation models  \citep{peng2025owsmv4improvingopen,li2025powsmphoneticopenwhisperstyle} and encoder-only models as well, where the additional task of phoneme recognition in \citet{li2025powsmphoneticopenwhisperstyle} may potentially induce even stronger cross-script alignment. Finally, in this work we study only the settings of X $\xrightarrow{}$ Cyrillic and X $\xrightarrow{}$ Latin transliteration, along with transliteration between traditional and simplified Chinese. Future work can expand this investigation to more transliteration directions.

\section*{Ethical Considerations}
All data used in this work are publicly available. While our experiments focus on understanding the representation of script information in speech foundation models, we acknowledge that activation steering and probing may be misapplied to either generate or detect content associated with certain demographic information. Our study is intended towards the goal of interpretable speech recognition, and we encourage its use towards building systems that embrace empirical linguistic diversity and the self-expression of its speakers.

\section*{Use of AI Assistants}
The authors acknowledge the usage of ChatGPT as an assistant tool in part of the source code’s development, in assisting the creation of figures, and in enhancing the coherence of parts of the manuscript.

\bibliography{custom}

\appendix

\section{Appendix}
\label{sec:appendix}

\FloatBarrier

\begin{table}
\centering
\small
\begin{tabular}{ll}
\toprule
\textbf{Script} & \textbf{Prompt} \\
\midrule
Simplified Chinese &
\begin{CJK}{UTF8}{gbsn}
这是一句普通话
\end{CJK} \\
Traditional Chinese & 
\begin{CJK}{UTF8}{bsmi}
這是一句普通話
\end{CJK} \\
Serbian (Latin) & Ovo je srpska rečenica \\
Serbian (Cyrillic) & \foreignlanguage{russian}{Ово је српска реченица} \\
\bottomrule
\end{tabular}
\caption{Script confusion prompts. All prompts have the meaning of "this is a sentence in X", where X is the actual language.}
\label{tab:script-confusion-prompts}
\end{table}

\begin{table}
\centering
\small
\label{tab:script-confusion-prompts}
\begin{tabular}{ll}
\toprule
\textbf{Language} & \textbf{Prompt} \\
\midrule
Russian & eto russkoye predlozheniye\\
Greek & auti einai mia elliniki protasi \\
Hindi & yah ek hindi vaakya hai\\
Korean & igeoseun hangugeo munjangibnida\\
Japanese & kore wa nihongo no bun desu\\
\bottomrule
\end{tabular}
\caption{Romanization prompts. All prompts have the meaning of "this is a sentence in X", where X is the actual language.}
\end{table}

\begin{table}
\centering
\small
\label{tab:script-confusion-prompts}
\begin{tabular}{ll}
\toprule
\textbf{Language} & \textbf{Prompt} \\
\midrule
Italian & \foreignlanguage{russian}{куэста э уна фразе итальяна}\\
Greek &  \foreignlanguage{russian}{афти инэ мия эллиники протаси}\\
Hindi & \foreignlanguage{russian}{йе эк хинди вакья хэ}\\
Korean & \foreignlanguage{russian}{игосын хангуго мунджанимнида}\\
Japanese & \foreignlanguage{russian}{корэ ва нихонго но бун десу}\\
\bottomrule
\end{tabular}
\caption{Cyrillization prompts. All prompts have the meaning of "this is a sentence in X", where X is the actual language.}
\end{table}

\begin{table}
\centering
\resizebox{0.5\textwidth}{!}{%
\begin{tabular}{llccccccc}
\hline\hline
Script & Method &
\multicolumn{5}{c}{Model Size} \\
\cline{3-9}
 & & tiny & base & small & medium & large  & v2  & v3 \\
\hline\hline

\multirow{3}{*}{sr-latn}
 & no-prompt            & \textbf{0.66} & \textbf{0.81} & \textbf{0.9} & 0.63 & 0.93 & 0.81 & 0.93 \\
 & prompt               & 0.64 & \textbf{0.81} & \textbf{0.9} & \textbf{0.93} & 0.95 & 0.95 & 0.95 \\
   & steer              & 0.64 & 0.8  & 0.89 & \textbf{0.93} & \textbf{0.96} & \textbf{0.96} & \textbf{0.96} \\
\hline

\multirow{3}{*}{sr--cyrl}
 & no-prompt           & 0.1 & 0.12 & 0.14 & 0.44 & 0.18 & 0.3 & 0.18 \\
 & prompt               & 0.12 & 0.21 & 0.52 & \textbf{0.91} & 0.82 & 0.77 & 0.82 \\
& steer   & \textbf{0.54} & \textbf{0.72} & \textbf{0.83} & 0.84 & \textbf{0.93} & \textbf{0.94} & \textbf{0.93} \\
\hline\hline

\multirow{3}{*}{zh--sim}
& no-prompt           & 0.6 & 0.67 & 0.79 & 0.87 & \textbf{0.93} & 0.85 & \textbf{0.93} \\
 & prompt               & \textbf{0.7} & \textbf{0.78} & \textbf{0.87} & \textbf{0.91} & \textbf{0.93} & \textbf{0.93} & \textbf{0.93} \\
 & steer     & 0.69 & \textbf{0.78} & \textbf{0.87} & \textbf{0.91} & \textbf{0.93} & \textbf{0.93} & \textbf{0.93} \\
\hline

\multirow{3}{*}{zh--trad}
& no-prompt           & 0.63 & 0.69 & 0.69 & 0.69 & 0.65 & 0.73 & 0.65 \\
 & prompt               & \textbf{0.71} & \textbf{0.79} & \textbf{0.86} & \textbf{0.9} & 0.79 & \textbf{0.91} & 0.79 \\
 & steer     & \textbf{0.71} & \textbf{0.79} & \textbf{0.86} & 0.89 & \textbf{0.89} & \textbf{0.91} & \textbf{0.89} \\
\hline\hline

\end{tabular}
}%
\caption{Script confusion across Whisper models.}
\label{tab:script_confusion}
\end{table}

\begin{table}
\centering
\resizebox{0.5\textwidth}{!}{%
\begin{tabular}{llccccccc}
\hline\hline
Script & Method &
\multicolumn{5}{c}{Model Size} \\
\cline{3-7}
 & & tiny & base & small & medium & large & v2 & v3 \\
\hline\hline

\multirow{3}{*}{sr-latn}
 & no-prompt            & \textbf{0.66} & \textbf{0.81} & \textbf{0.9} & 0.63 & 0.93 & 0.81 & 0.93 \\
 & prompt               & 0.64 & \textbf{0.81} & \textbf{0.9} & \textbf{0.93} & 0.95 & 0.95 & 0.95 \\
 & steer     & 0.63 & 0.8 & 0.89 & \textbf{0.93} & \textbf{0.96} & \textbf{0.96} &  \textbf{0.96} \\
\hline

\multirow{3}{*}{sr--cyrl}
 & no-prompt           & 0.1 & 0.12 & 0.14 & 0.44 & 0.18 & 0.3 & 0.18 \\
 & prompt               & 0.12 & 0.21 & 0.52 & \textbf{0.91} & 0.82 & 0.77 & 0.82 \\
 & steer     & \textbf{0.5} & \textbf{0.71} & \textbf{0.83} & 0.84 & \textbf{0.93} & \textbf{0.94} &  \textbf{0.93}\\
\hline\hline

\multirow{3}{*}{zh--sim}
& no-prompt           & 0.6 & 0.67 & 0.79 & 0.87 & \textbf{0.93} & 0.85 & \textbf{0.93} \\
 & prompt               & \textbf{0.7} & \textbf{0.78} & \textbf{0.87} & \textbf{0.91} & \textbf{0.93} & \textbf{0.93} & \textbf{0.93} \\
 & steer     & \textbf{0.7} & \textbf{0.78} & \textbf{0.87} & \textbf{0.91} & \textbf{0.93} & 0.91 & \textbf{0.93} \\
\hline

\multirow{3}{*}{zh--trad}
& no-prompt           & 0.63 & 0.69 & 0.69 & 0.69 & 0.65 & 0.73 & 0.65 \\
 & prompt               & \textbf{0.71} & \textbf{0.79} & \textbf{0.86} & \textbf{0.9} & 0.79 & 0.91 & 0.79 \\
 & steer     & \textbf{0.71} & \textbf{0.79} & \textbf{0.86} & \textbf{0.9} & \textbf{0.89} & \textbf{0.93} & \textbf{0.89} \\
\hline\hline
\end{tabular}
}%
\caption{One-shot script steering results across Whisper models.}
\label{tab:script_confusion_one_shot}
\end{table}

\FloatBarrier

\begin{table}
\small
\centering
\begin{tabular}{llcc}
\hline\hline
Script & Method & Romanization & Cyrillization \\
\hline\hline
\multirow{4}{*}{Hindi}
 & no-prompt        & 0.01 &  0.00\\
 & prompt           & 0.02 &  0.01\\
 & zero-shot        & 0.69 & \textbf{0.19}\\
 & pseudo-label     & \textbf{0.71} & 0.17\\
\hline
\multirow{4}{*}{Greek}
 & no-prompt        & 0.02 &  0.01\\
 & prompt           & 0.02 & 0.01 \\
 & zero-shot        & 0.53 &  \textbf{0.15}\\
 & pseudo-label     & \textbf{0.73} &  0.09\\
\hline
\multirow{4}{*}{Japanese}
 & no-prompt        & 0.01 &  0.01\\
 & prompt           & 0.01 &  0.01\\
 & zero-shot        & 0.28 &  \textbf{0.05}\\
 & pseudo-label     & \textbf{0.62} &  0.04\\
\hline
\multirow{4}{*}{Korean}
 & no-prompt        & 0.01 &  0.01\\
 & prompt           & 0.01 &  0.01\\
 & zero-shot        & 0.2 &  0.06\\
 & pseudo-label     & \textbf{0.51} &  \textbf{0.17}\\
\hline
\multirow{4}{*}{Russian}
 & no-prompt        & 0.01 & \textit{NA}  \\
 & prompt           & 0.01 & \textit{NA}  \\
 & zero-shot        & 0.67 & \textit{NA}  \\
 & pseudo-label     & \textbf{0.73} & \textit{NA}  \\
\hline
\multirow{4}{*}{Italian}
 & no-prompt        & \textit{NA}  & 0.01 \\
 & prompt           & \textit{NA}  & 0.01 \\
 & zero-shot        & \textit{NA}  & \textbf{0.43} \\
 & pseudo-label     & \textit{NA}  & 0.15 \\
\hline\hline
\end{tabular}
\caption{Romanization and cyrillization performance for Whisper large-v2.}
\label{tab:v2_merged}
\end{table}

\begin{table}
\small
\centering
\begin{tabular}{llccc}
\hline\hline
Language & Train & Validation & Test  \\
\hline\hline

Mandarin & 3246 & 409 &  945  \\
\hline

Serbian & 2944 & 290 &  700 \\
\hline

Russian & 2562 & 356 & 775  \\
\hline

Italian & 3030 & 391 &  865  \\
\hline

Hindi & 2120 & 239 & 418  \\
\hline

Greek & 3215& 271 & 650\\
\hline

Japanese & 2292&  266&  650 \\
\hline

Korean & 2307 & 226 &  382 \\
\hline\hline

\end{tabular}
\caption{Statistics of the train, validation, and test data in the FLEURS \cite{fleurs} dataset we employ.}
\label{tab:data_stats}
\end{table}

\begin{table}
\small
\centering
\begin{tabular}{llcc}
\hline\hline
Script & Method & Romanization & Cyrillization \\
\hline\hline

\multirow{2}{*}{Hindi}
 & zero-shot    & 0.2 & 0.3 \\
 & pseudo-label & 0.1 & 0.1 \\
\hline

\multirow{2}{*}{Greek}
 & zero-shot    & 0.3 & 0.3 \\
 & pseudo-label & 0.1 & 0.1 \\
\hline

\multirow{2}{*}{Japanese}
 & zero-shot    & 0.3 & 0.3 \\
 & pseudo-label & 0.2 & 0.5 \\
\hline

\multirow{2}{*}{Korean}
 & zero-shot    & 0.3 & 0.3 \\
 & pseudo-label & 0.1 & 0.5 \\
\hline

\multirow{2}{*}{Russian}
 & zero-shot    & 0.2 & \textit{NA} \\
 & pseudo-label & 0.1 & \textit{NA} \\
\hline

\multirow{2}{*}{Italian}
 & zero-shot    & \textit{NA} & 0.2 \\
 & pseudo-label & \textit{NA} & 0.1 \\
\hline\hline

\end{tabular}
\caption{Best sigmas for transliteration.}
\label{tab:sigmas}
\end{table}

\end{document}